\documentclass[11pt,a4paper]{article}
\usepackage[hyperref]{emnlp-ijcnlp-2019}

\usepackage{times}
\usepackage{latexsym}
\usepackage{hyperref}
\usepackage{graphicx}
\usepackage{multirow}
\usepackage{booktabs}
\usepackage{xcolor}

\usepackage{subcaption}
\usepackage{url}

\definecolor{darkblue}{rgb}{0, 0, 0.5}
\hypersetup{
    colorlinks=true,       
    linkcolor=darkblue,          
    citecolor=darkblue,        
    filecolor=darkblue,      
    urlcolor=darkblue          
}
\aclfinalcopy 

\newcommand{\cls}{{\textbf{[CLS]}}}

\title{Emergent Properties of Finetuned Language Representation Models}

\author{Alexandre Matton${}^{\dagger}$ \\
  Stanford University \& Twilio AI\\
  {\tt amatton@stanford.edu} \\\And
  Luke de Oliveira \\
  Twilio AI \\
  {\tt ldeoliveira@twilio.com} \\}

\date{}

\begin{document}

\maketitle
\begin{abstract}

{\let\thefootnote\relax\footnote{{${}^{\dagger}$Work completed as part of an internship at Twilio AI.}}}

\setcounter{footnote}{0}

Large, self-supervised transformer-based language representation models have recently received significant amounts of attention, and have produced state-of-the-art results across a variety of tasks simply by scaling up pre-training on larger and larger corpora. Such models usually produce high dimensional vectors, on top of which additional task-specific layers and architectural modifications are added to adapt them to specific downstream tasks. Though there exists ample evidence that such models work well, we aim to understand \emph{what happens} when they work well. We analyze the redundancy and location of information contained in output vectors for one such language representation model -- BERT. 
We show empirical evidence that the {\cls} embedding in BERT contains highly redundant information, and can be compressed with minimal loss of accuracy, especially for finetuned models, dovetailing into open threads in the field about the role of over-parameterization in learning. We also shed light on the existence of specific output dimensions which alone give very competitive results when compared to using all dimensions of output vectors.

\end{abstract}

\section{Introduction \& Related Work}

The recent emergence of successful large-scale language representation models (c.f.,~\citet{Devlin2018BERTPO,radford2019language,yang2019xlnet,liu2019roberta,shoeybi2019megatron,keskar2019ctrl}) has led to an explosion of \emph{self-supervised} language representation models trained on massive corpora of internet text. Even though these large models are trained on self-supervision\footnote{There is a larger debate, which we do not intend to participate in, as to whether or not such tasks are ``self-supervised'', ``unsupervised'', ``supervised'', or something in-between.} tasks such as Next-Sentence Prediction~\citep{Devlin2018BERTPO}, Masked-Language Modeling~\citep{Devlin2018BERTPO}, and vanilla next-token-prediction language models, such models perform well on tasks such as entailment or question answering for which they were not trained, whether in a zero-shot fashion (\citet{radford2019language,wang2019cross}, among others) or in a finetuned setting~\citep{Devlin2018BERTPO,radford2019language,yang2019xlnet,liu2019roberta}.

Although such models perform well, relatively little is known about why they work~\citep{hao2019visualizing}. Recent work has shown that properly tuning such models with nearly identical configuration and simply training on more data~\citep{liu2019roberta} leads to large gains in performance. Such models have also been shown to be highly compressible either through distillation~\citep{sanh2019distilbert,jiao2019tinybert} or quantization~\citep{zafrir2019q8bert,shen2019q}.

An open question remains as to whether or not such models \emph{have} to be over-parametrized~\citep{zhang2019identity} in order to perform well.~\citet{kovaleva2019revealing} and~\citet{hao2019visualizing}, among others, have speculated that models such as BERT~\citep{Devlin2018BERTPO} are \emph{vastly} over-parametrized for the downstream tasks on which they are finetuned. In this work, we provide empirical evidence in favor of this view, specifically for text classification.


\section{Experimental Setup}
\label{sec:experimental-setup}
We conduct all experiments on the task of fine-tuning our model for text classification -- in particular, we evaluate on three standard datasets for the aforementioned task; IMDB~\citep{maas2011learning}, AG-news~\citep{zhang2015character} and DBpedia~\citep{zhang2015character}. These datasets have very different numbers of examples, and final target space cardinality (see Table~\ref{dataset_sizes}). Running our analyses on these datasets provides an initial control as to whether the conclusions we draw are generalizable.

For simplicity, we fine-tune BERT with an identical experimental setup for each dataset. Each model is finetuned from the pretrained \emph{BERT-base}~\citep{Devlin2018BERTPO} model configuration 
which uses a hidden size of 768 with 12 transformer blocks and 12 attention heads. To adapt it for classification, we follow~\citet{Devlin2018BERTPO} and add a simple linear layer on top of the {\cls} token embedding. 

We follow~\citet{finetune_bert} for the selection of hyperparameters. We utilize a batch-size of 32 via gradient accumulation. We use the AdamW optimizer \`{a} la~\citet{loshchilov2017fixing} with $\beta_1=0.9$ and $\beta_2=0.999$, and combine this with slanted triangular learning rate decay~\citep{ulmfit}, with a cycle of 4 epochs, warm-up proportion of 0.1 and minimum learning rate of 0. Moreover, we adapt the maximum learning rate to each layer. The last layer is initialized with a maximum learning rate of $2 \times 10^{-5}$, and a decay of 0.95 is applied for each layer below (i.e. $LR_{n} = 0.95 \times LR_{n+1}$, where $n$ is the layer number). Anecdotally, we observed that using slanted triangular learning rates and decaying learning rates slightly improved results for all datasets. In addition, we use a dropout~\citep{srivastava2014dropout} value of $0.1$ post-{\cls}-token.

We respectively retained 10\%, 5\%, and 5\% of the training set of IMDB, AG-news, and DBpedia for early-stopping, and observed a convergence in under four epochs for all experiments. Our base-level fine-tuning results on the test sets are given in Table~\ref{Bert_results}.

We sanity-checked our initial results through~\url{https://paperswithcode.com/}. At the time this paper was written, our results place us in 9\textsuperscript{th} place for IMDB, 4\textsuperscript{th} place for AG-news and 2\textsuperscript{nd} place for DBpedia. Although ancillary from the primary argument of this work, this epitomizes how powerful these pretrained systems are: with only a few hours and for close to no cost, it is now possible to train models which get close to state-of-the-art results on most datasets.

\begin{table}[t!]
\small
\begin{center}
\begin{tabular}{lllc}
\hline \bf Dataset & \bf Training Set & \bf Test Set & \bf Output Classes \\ \hline
IMDB & 25,000 & 25,000 & 2\\
AG-news & 120,000 & 7600 & 4\\
DBpedia & 560,000 & 70,000 & 14\\
\hline
\end{tabular}
\end{center}
\caption{\label{dataset_sizes} Example and output cardinality of datasets used in experiments as elaborated upon in Section~\ref{sec:experimental-setup}.}
\end{table}

\begin{table}[t!]
\small
\begin{center}
\begin{tabular}{llll}
\hline \bf Dataset & \bf Pretrained & \bf Finetuned & \bf SOTA\\ \hline
IMDB & 88.0\% & 93.7\% & 97.4\%\\
AG-news & 90.4\% & 94.7\% & 95.5\%\\
DBpedia & 99.02\% & 99.33\% & 99.38\%\\
\hline
\end{tabular}
\end{center}
\caption{\label{Bert_results} Test-set performance of finetuned BERT over each experimental setup.}
\end{table}

\section{Results and Analyses}

In the following exposition, \emph{``pretrained models''} refers to models that have not been fully finetuned, where a linear classifier is trained on top of the pretrained embedding from the {\cls} token, but the BERT model itself is frozen. We refer to a \emph{``finetuned model''} as having been fully end-to-end finetuned on the relevant dataset and task.

Several experiments were conducted to inspect and understand the outputs of BERT-based models, in both pretrained-only and finetuned states. Most results and observations are consistent through all three datasets, which points towards the generalizability of the emergent properties derived from our studies toward other domains.

\subsection{Dimension Reduction via Principal Component Analysis}

To begin to understand how information is stored and represented in BERT vectors, we begin by analyzing model outputs through Principal Component Analysis (PCA), in which we seek to decompose a set of such vectors into an orthogonal basis set such that we may begin to understand dimensions of maximal variance in BERT vectors.

\subsubsection{Patterns in the output of pretrained BERT}

Do dimensions with maximal variance generalize across domains for text classification? An answer to this question points to a question of "how multi-task" components of BERT representations are. To investigate, we probe BERT using a general-domain PCA model, and attempt to understand how compress-able BERT representations are when carried cross-domain.

{\cls} token embeddings on a random sample of 1M sentences from a cleaned, English-language Wikipedia dump were computed, with a PCA model subsequently trained atop these embeddings. Post-training, this BERT + PCA combination was trained with a linear classifier on each of the datasets in our experimental setup. 

\begin{figure}
  \centering
  \includegraphics[width=.7\linewidth]{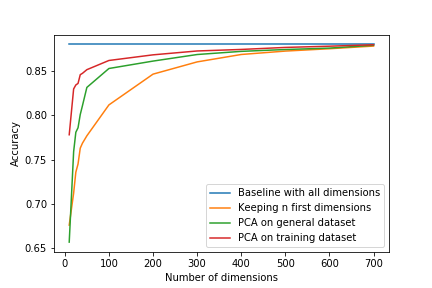}
  \caption{Accuracy of pretrained BERT on IMDB using PCA}
  \label{acc_BERT_pretrained_IMDB_with_PCA}
\end{figure}

\begin{figure}
  \centering
  \includegraphics[width=.7\linewidth]{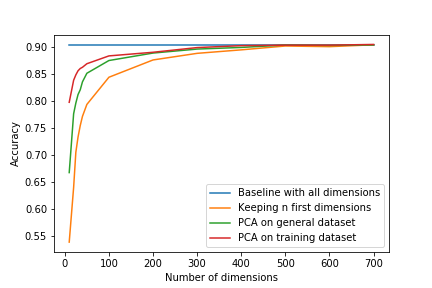}
  \caption{Accuracy of pretrained BERT on AG-news using PCA}
  \label{acc_BERT_pretrained_AG-news_with_PCA}
\end{figure}

\begin{figure}
  \centering
  \includegraphics[width=.7\linewidth]{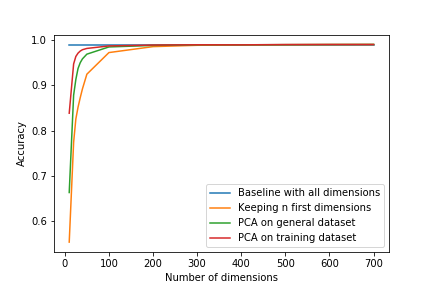}
  \caption{Accuracy of pretrained BERT on DBpedia using PCA}
  \label{acc_BERT_pretrained_DBpedia_with_PCA}
\end{figure}

We compare the performance under of this process with two alternatives -- the case where the PCA model is trained on the dataset of interest (removing the out-of-domain component), and the case where a random number of BERT dimensions is selected. We show the behavior under these three scenarios for IMDB in Figure~\ref{acc_BERT_pretrained_IMDB_with_PCA}, for AG-News in Figure~\ref{acc_BERT_pretrained_AG-news_with_PCA}, and for DBPedia in Figure~\ref{acc_BERT_pretrained_DBpedia_with_PCA}.

The behavior in Figures~\ref{acc_BERT_pretrained_IMDB_with_PCA},~\ref{acc_BERT_pretrained_AG-news_with_PCA}, and~\ref{acc_BERT_pretrained_DBpedia_with_PCA} clearly demonstrate that there indeed is generalizable structure and information in BERT outputs which makes a large portion of output features redundant. Although performance clearly degrades when the PCA model is trained out-of-domain versus in-domain, the gap between results is smaller than the gap when compared to randomly selected dimensions. 

An interesting anecdotal result is that PCA does not require a large sample to reach full capacity, as evaluated by downstream performance as a feature set on a given task. We initially trained PCA on around 1M embeddings, but after manual analysis, it was determined that around 32,000 embeddings are sufficient to obtain performance on par with the full set of 1M. 

\subsubsection{Patterns in the output of finetuned BERT}

We empirically observe that when BERT is finetuned, the information is \emph{extremely} compressed into a small number of dimensions in the {\cls} token embedding -- surprisingly good performance is obtained with only a handful of dimensions. With 5 principal components obtained from the general PCA, scores are only 0.004\% (IMDB) / 0.35\% (AG-news) / 1.05\% (DBpedia) percentage points away from models trained atop of all principal components. With 25 principal components, it is the percentage point gap for accuracy drops to 0.02\% for DBpedia, and significantly less for others. Hence, output information can be drastically compressed.

\subsubsection{Explained variance}

Principal Component Analysis admits the useful property of maximizing the variance of the projected data across each component, subject to maintaining orthogonality to existing components. We utilize this fact, as is commonly done, to examine the percentage of observed variance encoded in each projected axis. In training PCA on the general Wikipedia dataset, we observed shared phenomena across all three datasets.

We observe that the first principal axes become more important when the model is finetuned, i.e they explain significantly more variance than their counter-parts in the pretrained models. However, the number of such principal axes is low. Surprisingly, the point at which we encounter the first principal component that explains less variance in the finetuned model than the pretrained model roughly corresponds to the number of classes contained in the task at hand.

\begin{figure}
  \centering
    \includegraphics[width=.7\linewidth]{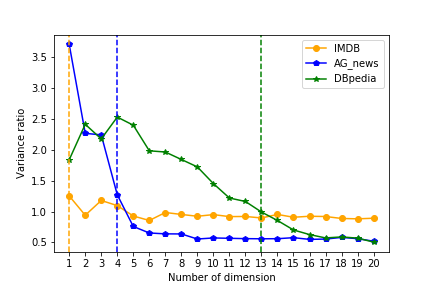}
    \caption{Variance ratios. The vertical lines correspond to the last index for which the first variance ratios are greater than 1. We define a variance ratio to be the ratio of the variance explained for the $i$\textsuperscript{th} component in the finetuned model relative to the pretrained model.}
    \label{variance_ratios}
\end{figure}

Define the \textit{i\textsuperscript{th} variance ratio} to be the percentage of explained variance of $i$\textsuperscript{th} principal component of the finetuned model divided by $i$\textsuperscript{th} principal component in the pretrained model. We display the variance ratios for the first 20 principal components across the three datasets under consideration in Figure~\ref{variance_ratios}.

For AG-news, we note that the first four axes are significantly more important in the finetuned model than in the pretrained-only model, with variance ratios between 1.3 and 3.7, before dropping to a variance ratio of 0.76 in the fifth principal component. For DBpedia, the first 13 variance ratios are greater or equal to 1, and for IMDB, the first variance ratio is 1.25, and drops to below 1 for the second dimension.\footnote{As a reminder, AG-news has 4 output categories, DBPedia has 14 output categories, and IMDB has 2 output categories (See Table~\ref{dataset_sizes}).}
    
We hypothesize, without substantive testing, that the BERT embeddings corresponding to {\cls} tokens develop a natural dimensionality close to the number of natural categories of each dataset (c.f. Table~\ref{dataset_sizes}). This is consistent with the intuition that BERT is vastly overparameterized for many of the tasks at hand\footnote{We suspect this is because the standard text classification chosen here tend to require little-to-no reasoning.}.

This hypothesis agrees with the ``Natural'' column of Table~\ref{sentiment_neuron_score}, in which we look for subsets of embedding dimensions that provide similar results to our full system, without doing any PCA. We elaborate upon this in Sections~\ref{sentimental-neuron-pretrained} and~\ref{sentimental-neuron-finetuned}.

\subsection{Does BERT have salient neurons?}

\begin{table*}[t!]
\centering
\begin{tabular}{@{}lrlllllll@{}}
\cmidrule(l){3-9}
        &              & \multicolumn{3}{c}{Pretrained}        & \multicolumn{4}{c}{Finetuned}       \\ \cmidrule(l{10pt}r{10pt}){2-5} \cmidrule(l{10pt}r{10pt}){6-9}
Dataset & Random       & All     & 1       & 5       & All     & 1      & 5      & Natural \\ \midrule
IMDB    & 50 \% (2)    & 88.0\%  & 67.0\%  & 77.4\%  & 93.7\%  & 93.7\% & 93.8\% & 93.7\%    \\
AG-news & 25 \% (4)    & 90.4\%  & 43.5\%  & 70.3\%  & 94.7\%  & 83.6\% & 94.3\% & 94.3\%    \\
DBpedia & 7.14 \% (14) & 99.02\% & 17.67\% & 53.67\% & 99.33\% & 60.9\% & 99.0\% & 99.2\%    \\ \bottomrule
\end{tabular}
\caption{\label{sentiment_neuron_score} Accuracy of a simple softmax classifier on top of the {\cls} BERT embedding while: keeping all dimensions, keeping only the best dimension, and keeping the five best dimensions. Leftmost column corresponds to the accuracy of random guesses.}
\end{table*}

We now investigate how individual dimensions of BERT {\cls} embeddings learn to store information, removing the explicit cumulative variance-maximizing projections of PCA in the previous section. To do this, we explicitly select output dimensions that are most useful for each specific task. We refer to these dimensions as \emph{salient neurons}, a slight nod towards sentiment neurons referenced in~\citet{sentiment_neuron}.

\subsubsection{Salient neurons in pretrained BERT}
\label{sentimental-neuron-pretrained}

We first begin by investigating, in a similar spirit to~\citet{sentiment_neuron}, which dimensions or neurons from the pretrained BERT model directly exhibit useful information for our three datasets. To select the best neuron for a given classification task, we perform 5-fold cross-validation on the train set and select the individual dimension with the best mean accuracy score over the folds. 

As the original number of dimensions to search over is high (768 for the BERT-base model), selecting the subset of size $n$ which maximizes our cross-validated accuracy score suffers from combinatorial explosion -- therefore, we proceed in a greedy fashion, choosing new dimensions to be added to existing ones by cross-validation\footnote{An interesting extension for future work is to understand whether a more optimal search for hidden unit combinations leads to better performance than the results shown in Table~\ref{sentiment_neuron_score}}.

Results for the pretrained-only model are presented in the left part of Table~\ref{sentiment_neuron_score}. As is evident, certain neurons encode information \emph{with no finetuning} that are useful for the classification tasks outlined in Section~\ref{sec:experimental-setup}, even though the model has not received explicit signal for these tasks before. This hints towards validation that for the auxiliary tasks that comprise BERT (i.e. masked words and sentence prediction), the model has learned to embed information which encapsulates sentiment and other topical categorizations. As is clear in Table~\ref{sentiment_neuron_score}, however, results on selected pretrained subsets of embedding dimensions are still quite far from those obtained from finetuned models, exhibiting the value of domain alignment via finetuning.

\subsubsection{Salient neurons in finetuned BERT}
\label{sentimental-neuron-finetuned}

\begin{figure}
  \centering
  \includegraphics[width=.7\linewidth]{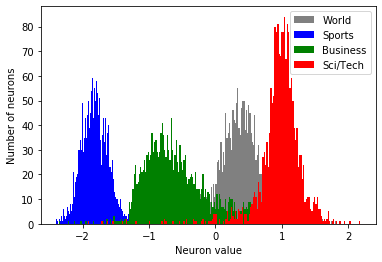}
  \caption{Best neuron activation for BERT finetuned on AG-news}
  \label{hist_neuron_value_wrt_label_bert_finetuned_agnews}
\end{figure}

With finetuned models, one would expect that salient neurons would manifest themselves more clearly. As shown in Table~\ref{sentiment_neuron_score}, this is confirmed in our experiments. These salient neurons are \emph{surprisingly effective}, in many cases removing the need for the fully finetuned classification layer. For the IMDB dataset, using a linear classification on even a single neuron from the {\cls}-token embedding provides equal performance to scores from models which utilize the full embedding. For AG-news and DBpedia this is less pronounced, primarily due to the number of classes contained in the given datasets. However, the best neuron manages to provide very strong separation between classes, and does much more than only separating one class from all the others, as shown in Figure~\ref{hist_neuron_value_wrt_label_bert_finetuned_agnews}, which clearly illustrates the power of a single dimensional representation from this neuron. An interesting anecdotal observation from our experiments is that the performance when selecting a number of neurons equal to the number of classes is approximately equal to the case where we utilize the entire embedding. This leads us to hypothesize that the over-parametrization of the finetuned BERT model for the tasks at hand results in embedding vectors lying on a significantly lower dimensional manifold than the full 768 dimensions, with this natural manifold tending towards the number of classes\footnote{A byproduct of the finetuning being explicitly for class separation.}.

A natural follow-up question to the previous analysis is whether or not salient neurons are unique -- put differently, are there many redundant salient neurons?

\begin{figure*}[t!]
    \centering
    \begin{subfigure}[t]{0.4\textwidth}
        \centering
        \includegraphics[width=0.99\textwidth]{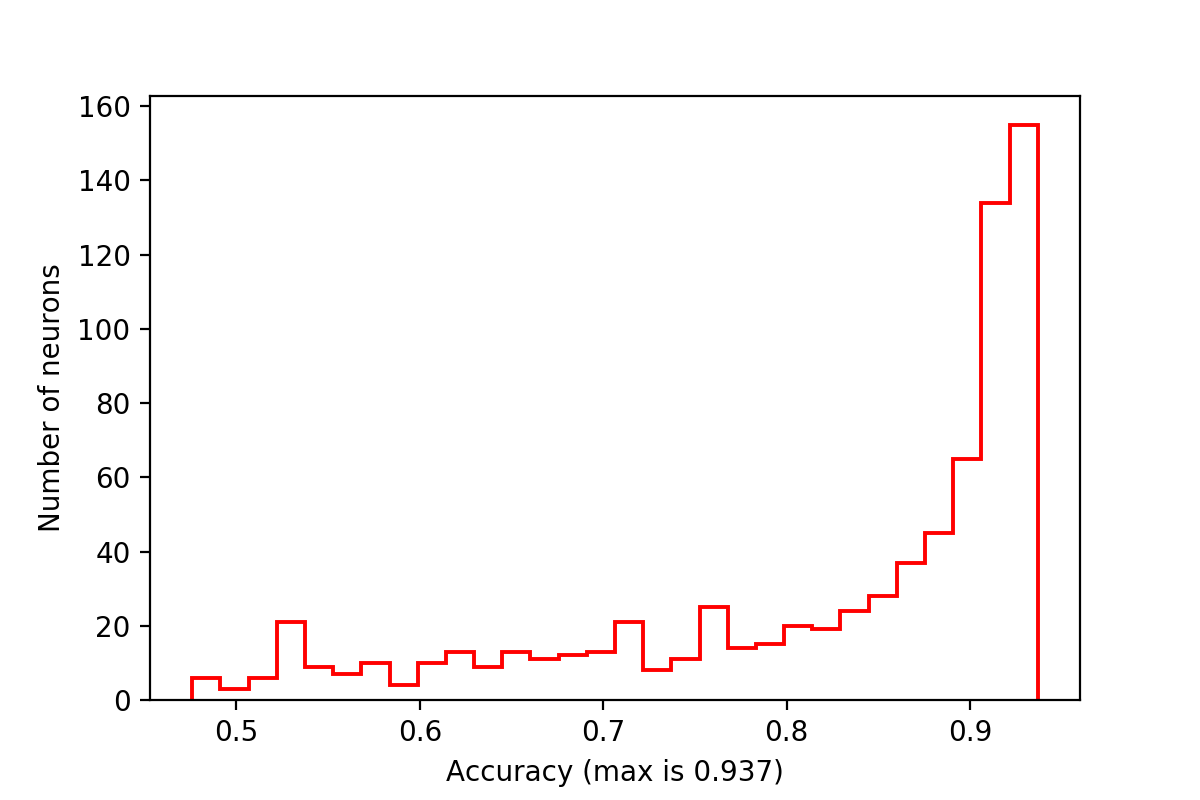}
        \caption{Histogram of IMDB accuracy}
        \label{hist_weight_repartition_logreg_BERT_finetuned_IMDB}
    \end{subfigure}%
    ~ 
    \begin{subfigure}[t]{0.4\textwidth}
        \centering
        \includegraphics[width=0.99\textwidth]{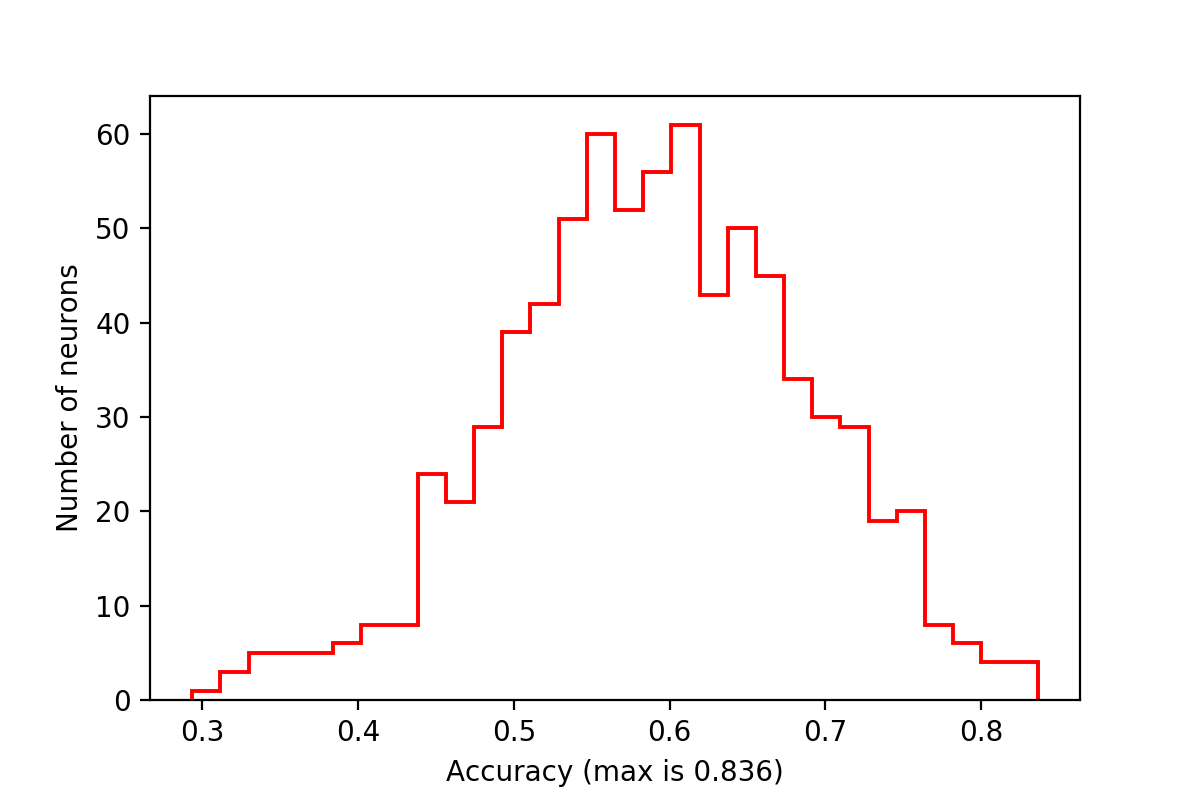}
        \caption{Histogram of AG-news accuracy}
        \label{hist_weight_repartition_logreg_BERT_finetuned_AG-news}
    \end{subfigure}
    ~ 
    \begin{subfigure}[t]{0.4\textwidth}
        \centering
        \includegraphics[width=0.99\textwidth]{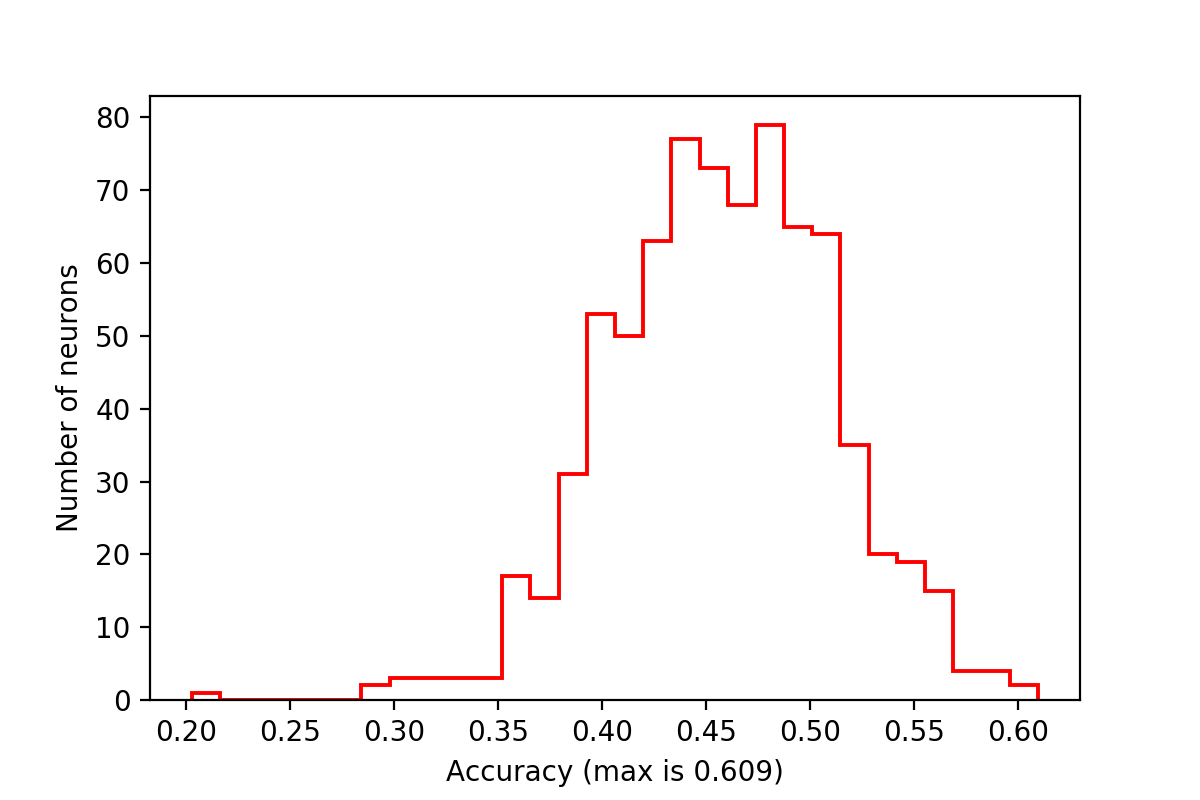}
        \caption{Histogram of DBPedia accuracy}
        \label{hist_weight_repartition_logreg_BERT_finetuned_DBPedia}
    \end{subfigure}
    \caption{Histograms of accuracy scores when selecting individual neurons from the 768-dimensional {\cls} embeddings from finetuned models on individual datasets.}
    \label{figure:acc-histograms}
\end{figure*}

To investigate, we consider the empirical accuracy distribution of individual classifiers trained on single elements of the {\cls} token embedding. In Figure~\ref{figure:acc-histograms}, we display these empirical distributions for IMDB (Fig.~\ref{hist_weight_repartition_logreg_BERT_finetuned_IMDB}), AG-News (Fig.~\ref{hist_weight_repartition_logreg_BERT_finetuned_AG-news}), and DBPedia (Fig.~\ref{hist_weight_repartition_logreg_BERT_finetuned_DBPedia}). As is evident, the distribution of accuracy scores is quite dense, i.e., there are no gaps in how individual neurons perform. Interestingly, the number of salient neurons that are \emph{highly performant} is quite low, though there does seem to be some redundant information. For example, on the IMDB dataset, around 7\% of neurons result in an accuracy greater than 0.93, but only a single neuron gives an accuracy score greater than 0.936\footnote{As a reminder, neurons are evaluated on the test set here, but when selecting neurons for inclusion in Table~\ref{sentiment_neuron_score}, selection occurs via cross-validation.}. 

\section{Discussion and Future Work}

In this work, we show empirical evidence that both the pretrained and finetuned representations in a BERT {\cls}-token embedding contain significant amounts of redundancy, and, in the case of finetuning, exhibit low dimensional manifold structure intimately related to the problem at hand. We conclude, through experimental investigation, that BERT is vastly over-parametrized for standard text classification tasks, echoing~\citet{sanh2019distilbert}.

One major consequence is that it is possible to reduce the dimension of BERT {\cls}-token output to a very small number of dimensions and lose almost no accuracy. This can be useful in settings where one needs to store a many such embeddings, such as text retrieval, or if a downstream task requires sending the embeddings through a network. In addition, we suspect that designing connective patterns between the multi-head attention components and the final {\cls}-token vectors that can take advantage of sparsity in the vector output may be useful for pruning-aware training.

As a field, natural language processing is currently under a deluge of transformer-based language representation models, where each one seems to be slightly better than all that came before. A key dimension of future work would be to run the same experiments on these new systems. We speculate that since their architectures are very similar, results are likely to be close.

Finally, our study exclusively concerned standard classification tasks. We intend to adapt this study to a wider range of tasks to determine if such strong evidence of over-parametrization exists across language understanding benchmarks.

\section*{Acknowledgments}

The authors would like to acknowledge useful conversations with Ian Lane, Jungsuk Kim, Osman Ba\c{s}kaya, Richie Feng, Alfredo L\'{a}inez, Warren Green, Umair Akeel, and the entire Twilio AI team.

\bibliography{paper_bibliography}
\bibliographystyle{acl_natbib}

\end{document}